\documentclass[10pt]{IEEEtran}

\usepackage{enumerate} 
\usepackage{geometry}
\usepackage{graphicx,amsmath,amsfonts}
\usepackage{url}

\usepackage{tabularx}
\usepackage{array, booktabs}
\usepackage{graphicx,amsmath,amsfonts}
\usepackage{wrapfig}
\usepackage{geometry}
\usepackage{subfig} 
\usepackage{caption}
\usepackage{epstopdf}
\usepackage{cite}
\usepackage{setspace}
\usepackage{graphics}
\usepackage{placeins}
\usepackage{float}
\restylefloat{table}
\begin{document}

\title{Predicting all-cause Hospital Readmissions from Medical Claims data of Hospitalised Patients}

\author{\IEEEauthorblockN{Avinash Kadimisetty\IEEEauthorrefmark{1},
Arun Rajagopalan\IEEEauthorrefmark{2} and 
Vijendra SK\IEEEauthorrefmark{3} \\}
\IEEEauthorblockA{Evive Software Analytics Pvt. Ltd.,
	Bengaluru, Karnataka, India\\
Email: \IEEEauthorrefmark{1}avinash.k@goevive.com,
\IEEEauthorrefmark{2}arun.rajagopalan@goevive.com,
\IEEEauthorrefmark{3}vijendra@goevive.com}}

\maketitle
\begin{abstract}
		Reducing preventable hospital readmissions is a national priority for payers, providers, and policymakers seeking to improve health care and lower costs. The rate of readmission is being used as a benchmark to determine the quality of healthcare provided by the hospitals. In this project, we have used machine learning techniques like Logistic Regression, Random Forest and Support Vector Machines to analyze the health claims data and identify demographic and medical factors that play a crucial role in predicting all-cause readmissions. As the health claims data is high dimensional, we have used Principal Component Analysis as a dimension reduction technique and used the results for building regression models. We compared and evaluated these models based on the Area Under Curve (AUC) metric. Random Forest model gave the highest performance followed by Logistic Regression and Support Vector Machine models. These models can be used to identify the crucial factors causing readmissions and help identify patients to focus on to reduce the chances of readmission, ultimately bringing down the cost and increasing the quality of healthcare provided to the patients.
\end{abstract}


\section{Introduction}
More than a trillion dollars are being spent annually in the healthcare industry partly due to the latest technology not being used to its fullest in healthcare. Machine Learning techniques can have a huge impact in reducing the healthcare costs that are expected to increase in the coming years.  Patients' readmissions to the hospitals are one of the reasons for the increasing costs in healthcare. Readmissions often occur due to poor treatment provided to the patients, more specifically, they are often caused by premature discharges or communication breakdown between patients and the healthcare team while the patient is being discharged. 

Unplanned Hospital Readmission is defined as an unexpected readmission to the same hospital within 28 or 30 days of being discharged. However, the literature has widely used 30 days within the context of measurement of Hospital Readmissions. Unplanned Hospital Readmission rate is considered as a performance indicator to measure a hospital's quality of care. 

Unplanned readmissions cause a disruption to the normality of the patients’ lives and result in significant financial burden on the healthcare system. In the USA alone, it has been estimated that 20\% (7.8 million) of the hospital discharged patients were readmitted. These readmissions result in higher costs to taxpayers, costing as much as \$45 billion annually. Medicare, along with other healthcare payers, are concerned with the cost of unnecessary readmissions as Medicare alone spends roughly \$15 billion annually on repeat hospitalizations. Almost 76\% of the repeat admissions can be avoided by improving care before and after the patient is discharged. By decreasing these preventable repeat hospitalizations, overall productivity of the hospitals can improve considerably. In this study, we aimed at building predictive models to identify patients at high risk for readmissions. Preventive approaches can then be developed and applied to target the identified high-risk patients. 

\section{Literature Survey}
As per the Affordable Care Act of 2010, hospitals reimburse a partial amount to the patients readmitted to the hospital within 30 days from discharge~\cite{protection2010patient}. As these readmissions are costly and considered as an indication of poor quality, many studies have been performed to identify various factors that play crucial roles in predicting possible readmissions thereby alleviating revenue losses. As the hospital readmissions are driven by the nature of the population, a few studies involved deeper analysis to provide richer and nuanced explanation of readmissions~\cite{8290026}. In one of the previous analyses on 30 day hospital readmissions, medical conditions like Chronic Obstructive Pulmonary Disorder (COPD), Total Hip Arthroplasty (THA) etc were used primarily~\cite{7364123}. In this analysis, both parametric and non-parametric statistical models and machine learning techniques like Gradient Boosting, Neural Networks and Decision Trees were used and the primary metric used for performance evaluation was Area Under Curve (AUC). Another study was performed to predict 30-day hospital readmission in COPD patients~\cite{7897269}. This study, apart from the medical factors contributing towards a readmission, incorporated cost and proposed several methods to directly incorporate cost into the prediction. 

A few studies focussed on all-cause readmissions where predictive models were built with various types of predictors. These studies included fixed patient attributes such as morbidity burden, maternity and disability etc while most focussed on general patient attributes such as previous acute hospital stays, accumulated days, count of emergency department episodes etc~\cite{6680534}. One of the studies involving prediction of all cause readmissions involved prediction of post discharge death and analysed the deterioration of model performance in population having multiple admissions per patient~\cite{walraven2013predicting}. A few studies that used LACE index analysed its poor peformance in some older population~\cite{cotter2012predicting}. \nocite{*} 

\section{Data Description}
The data for this study has been obtained from various health insurance providers in the USA. The demographics data, medical claims data and the pharmacy claims data are collected from these providers. The fields present in each of the datasets are desribed in the following sections.

\subsubsection{Demographics Data}
Demographics data is the statistical data which describes the characteristics of a population such as Age, Gender, Income, Race, Education, Employement etc. For the purpose of this study Gender, Age, Ethnicity and Scheme Type are considered as the demographic features of a person. \tablename{~\ref{demographicsDataDictionary}}~describes each of the above mentioned fields.

\begin{table}[!htb]
	\begin{center}
	\centering
	\begin{tabular}{|l|l|}
		\hline
		Field & Description \\
		\hline
		Gender & Describes the sex of the person \\ & (M - Male or F - Female) \\
		\hline
		Age & Age is a numerical value \\
		\hline
		Ethnicity & Describes the race of the person \\ & and takes the values White, \\ & Asian, Hispanic and Black \\
		\hline
		Scheme Type & Describes the living area of the \\ & person and takes the values \\ & Large Central Metro, Large \\ & Fringe Metro, Medium Metro, \\ & Small Metro, Micropolitan,\\ & Noncore
		 \\
		\hline
	\end{tabular}
	\caption{Demographics Data Description Table}
	\label{demographicsDataDictionary}
	\end{center}
\end{table}

\subsubsection{Medical Claims Data}
Medical claims data is the information available in medical billing claims forms filled on behalf of a population. This information is gathered from the medical claims submitted by health care providers to the health insurers. The information obtained from these providers is at an individual claim level and consists of the following fields, Service Start Date, Service End Date, Primary Diagnosis Code, Other Diagnosis Codes, CPT Code. The description for each of these fields is given in \tablename{~\ref{medicalClaimsDictionary}}.

\begin{table}[!htb]
	\begin{center}
		\centering
		\begin{tabular}{|l|l|}
			\hline
			Field & Description \\
			\hline
			Service Start & The begin date of a \\ date & medical service \\
			\hline
			Service End & The end date of a \\ date & medical service \\
			\hline
			Primary Diagnosis & The ICD code for the \\ Code & primary disease \\
			\hline
			Other Diagnosis & The ICD codes for the \\ Codes & other diagnosed diseases \\
			\hline
			CPT Code & The CPT code of the \\&  procedures undergone  \\ & during the hospital visit \\
			\hline
		\end{tabular}
		\caption{Medical Claims Data Description Table}
		\label{medicalClaimsDictionary}
	\end{center}
\end{table}

\subsubsection{Pharmacy Claims Data}
The pharmacy claims data is obtained from the health insurers. This data contains the information related to the drugs prescribed by a doctor - name of the drug, quantity, prescription date, purchase date, price of the drugs etc. For the purpose of this study, Service Date (purchase date) and the NDC Code (a unique 10-digit numeric identifier assigned to each medication) are considered. The description of these two fields is given in \tablename{~\ref{pharmacyClaimsDictionary}}.

\begin{table}[!htb]
	\begin{center}
		\centering
		\begin{tabular}{|l|l|}
			\hline
			Field & Description \\
			\hline
			Service Date & The date on which the pharmacy \\ & drug was purchased. \\
			\hline
			NDC Code & The NDC code representing \\ & the drug \\
			\hline
		\end{tabular}
		\caption{Pharmacy Claims Data Description Table}
		\label{pharmacyClaimsDictionary}
	\end{center}
\end{table}

\section{Data Processing}
In this study, we have conducted predictive modelling at the episode-level. The target variable is whether a patient was readmitted within 30 days or not after being discharged. Hence the target variable is a binary variable. The objective of this experiment is to construct predictive models to determine how various factors impact a patient’s re-hospitalization. The process of finding an episode is explained in the following paragraphs. Let us consider the Medical Claims \tablename{~\ref{MedClaimsTable}}, Demographics \tablename{~\ref{demographicsTable}} and Pharmacy Claims Table~\tablename{~\ref{phClaimsTable}} for explaining the data processing steps.

\begin{table*}[h]
	\centering
	\begin{tabular}{|c|c|c|c|c|c|c|}
		\hline
		UserID & ClaimID & Service & Service &
		Primary & Other & CPT\\
		&&Start&End&Diagnosis&Diagnosis&Code\\
		&&Date&Date&Code&Codes\\
		\hline
		User1 & C1 & 2017-04-01 & 2017-04-01 & 682.50 & 786.50 & 99211\\
		\hline
		User1 & C2 & 2017-05-01 & 2017-05-03 & 70890 & 40201 & 99281\\
		\hline
		User1 & C3 & 2017-05-04 & 2017-05-08 & 041.12 & 09320 & 61000\\
		\hline
		User1 & C4 & 2017-05-21 & 2017-06-09 & 186.19 & 00000 & 99231\\
		\hline
		User1 & C5 & 2017-07-01 & 2017-07-03 & 37234 & 34200 & 99231\\
		\hline
		User2 & C6 & 2018-01-03 & 2018-01-08 & 78903 & 49001 & 99231\\
		\hline
		User2 & C7 & 2018-01-03 & 2018-01-15 & 995.29 & 00000 & 43888\\		
		\hline
	\end{tabular}
	\caption{Medical Claims Data Table}
	\label{MedClaimsTable}
\end{table*}

As the medical claims data is at an individual claim level (which can be observed from~\tablename{~\ref{MedClaimsTable}}) and not at the episode level, certain methods of identifying an admission were followed. An admission can be identified by the inpatient CPT codes (99231-99236, 99224-99226, 99281-99285, 99291-99292) and the discharge can be identified by 99238, 99239, 99217. But the discharge CPT codes are not used often and hence this method could not be used. The length of stay heuristic is followed here which states that, \textit{Two individual claims are grouped together as a part of an episode if the difference between the service end date of the first claim and the service start date of the second claim is less than 10 days} \cite{lengthofstay}.

After the admissions of each user are found, an admission is treated as a readmission if the difference between the previous admission and the current admission is less than or equal to 30 days. If a particular admission is a readmission, it is removed from the list of admissions and added to the readmission list. In our data this grouping resulted in 40,358 admissions of which 1,880 are readmissions resulting in 4.65\% readmission rate.

\begin{table*}[h]
	\centering
	\begin{tabular}{|c|c|c|c|c|}
		\hline
		UserID & ID & Service & Service & Re-\\
		&&Start&End&admission\\
		&&Date&Date&\\
		\hline
		User1 & A1 & 2017-05-01 & 2017-05-08 & YES\\
		\hline
		User1 & A2 & 2017-07-01 & 2017-07-03 & NO\\
		\hline
		User2 & A3 & 2018-01-03 & 2018-01-15 & NO\\
		\hline
	\end{tabular}
	\caption{Admissions Table}
	\label{admissionTable}
\end{table*}


In the medical claims \tablename{\ref{MedClaimsTable}}, claim with ClaimID C2 has the inpatient CPT code. Claims C2 and C3 are grouped together as the difference between service end date and start date of the two claims is less than 10 days. Since the difference between the dates for C3 and C4 is greater than 10 days and less than 30 days, C4 is considered to be a readmission. Similary, the claims are grouped for User2 as well. The admissions table after grouping the claims using the above mentioned process is shown in \tablename{~\ref{admissionTable}}.

The predictor variables considered for this study are Comorbidities, Demographics, Length of Stay, Medications during the admission, Number of previous admissions, Number of previous emergency department admissions, Admitting Diagnosis, Number of previous hospital visits, and Admission procedures. All the predictor variables have been derived from the available data and the extraction process of each feature is explained in the following paragraphs. To illustrate the process of extracting each feature, the admissions \tablename{~\ref{admissionTable}} is considered.

\begin{table}[H]
	\centering
	\begin{tabular}{|c|c|c|c|}
		\hline
		UserID & ClaimID & Service Date & NDC Code \\
		\hline
		User1 & P1 & 2017-05-05 & 0002759701 \\
		\hline
		User1 & P2 & 2017-05-07 & 5024204062 \\
		\hline
		User2 & P3 & 2018-01-04 & 6057541121\\
		\hline
	\end{tabular}
	\caption{Pharmacy Claims Data Table}
	\label{phClaimsTable}
\end{table}

\subsubsection{Comorbidities}
Comorbidity is the presence of one or more additional diseases co-occurring with a primary disease. For the purpose of this study we considered the following comorbidities - CHF, Valvular, PHTN, PVD, HTN, HTNcx, Paralysis, NeuroOther, Pulmonary, DM, DMcx, Hypothyroid, Renal, Liver, PUD, HIV, Lymphoma, Mets, Tumor, Rheumatic, Coagulopathy, Obesity, WeightLoss, FluidsLytes, BloodLoss, Anemia, Alcohol, Drugs, Psychoses, Depression. Each of these comorbidities map to a set of ICD codes. Based on the field "Other Diagnosis Codes" in the medical claims data the comorbidities during each admission are identified. For each admission in the \tablename{~\ref{admissionTable}}, the comorbidities present are shown in Table~\tablename{~\ref{comorbiditiesTable}}.

\begin{table}[H]
	\centering
	\begin{tabular}{|c|c|c|}
		\hline
		UserID & ID & Comorbidities\\
		\hline
		User1 & A1 & CHF, Valvular\\
		\hline
		User1 & A2 & Paralysis\\
		\hline
		User2 & A3 & Pulmonary\\
		\hline
	\end{tabular}
	\caption{Comorbidities in each admission}
	\label{comorbiditiesTable}
\end{table}

\subsubsection{Demographics}
The demographic features considered in this study include Gender, Age Group, Ethnicity, Income Level and Scheme Type. Age group is derived by discretizing the Age into five groups Touch [0-20), Millennials [20-37), GenX [37-49), Boomers [49-68) and Swing (68+). The demographic features of each user in the admissions \tablename{~\ref{admissionTable}} are shown in~\tablename{~\ref{demographicsTable}}.
\begin{table}[H]
	\centering
	\begin{tabular}{|l|l|l|l|l|}
		\hline
		UserID & Gender & Age & Ethn- & Scheme \\
		&&&city&\\
		\hline
		User1 & M & 25 & Asian & Large \\
		& & & & Central \\
		& & & & Metro\\
		\hline
		User2 & F & 35 & White & Medium\\
		& & & & Metro\\
		\hline
	\end{tabular}
	\caption{Demographics Data Table}
	\label{demographicsTable}
\end{table}

\subsubsection{Length of Stay}
The Length of Stay (LOS) of each admission is derived by subtracting the admission date (which is the service start date of the episode) from the discharge date (which is the service end date of the episode). This a numerical feature and can take values starting from 0. For each admission in the \tablename{~\ref{admissionTable}}, the length of stay is indicated in \tablename{~\ref{losTable}}.
\begin{table}[H]
	\centering
	\begin{tabular}{|c|c|c|}
		\hline
		UserID & ID & LOS \\
		\hline
		User1 & A1 & 8 days\\
		\hline
		User1 & A2 & 3 days\\
		\hline
		User2 & A3 & 13 days\\
		\hline
	\end{tabular}
	\caption{Length of stay for each admission}
	\label{losTable}
\end{table}

\subsubsection{Medications}
The NDC codes of the drugs taken during the admission are fetched from the pharmacy claims. These NDC codes are categorized into 100 groups using GPI level 2 categorization. For exampple, a drug with NDC Code 6057541121 will have 60 (first two digits) as its GPI level 2 category. These 100 categories take binary values.For each admission in the \tablename{~\ref{admissionTable}}, the medications taken are shown in \tablename{~\ref{medicationTable}}.
\begin{table}[!htb]
	\centering
	\begin{tabular}{|c|c|c|}
		\hline
		UserID & ID & Medications\\
		\hline
		User1 & A1 & 00, 50\\
		\hline
		User1 & A2 & None\\
		\hline
		User2 & A3 & 60\\
		\hline
	\end{tabular}
	\caption{Number of previous admissions}
	\label{medicationTable}
\end{table}

\subsubsection{Number of Previous Admissions}
Using the derived admissions data from the medical claims data, the number of previous admissions for each admission is derived by taking the count of admissions whose admission date is before the service start date of the current admission. This is a numerical feature and the value is a whole number. For each admission in the \tablename{~\ref{admissionTable}}, the number of previous admissions is shown in \tablename{~\ref{nopaTable}}.
\begin{table}[!htb]
	\centering
	\begin{tabular}{|c|c|c|}
		\hline
		UserID & ID & Number of \\ &&Previous Admissions\\
		\hline
		User1 & A1 & 0\\
		\hline
		User1 & A2 & 1\\
		\hline
		User2 & A3 & 0\\
		\hline
	\end{tabular}
	\caption{Number of previous admissions}
	\label{nopaTable}
\end{table}

\subsubsection{Number of Previous Emergency Department Admissions}
The emergency department admissions are identified by the set of CPT codes 99281-99285 and these admissions are filtered from the admission data. The number of previous emergency admissions is derived by taking the count of emergency admissions whose admission date is before the service start date of the current admission. This is a numerical feature and the value is a whole number. For each admission in the \tablename{~\ref{admissionTable}}, the number of previous emergency department admissions is shown in \tablename{~\ref{nopedaTable}.
\begin{table}[!htb]
	\centering
	\begin{tabular}{|c|c|c|}
		\hline
		UserID & ID & Number of previous\\
		& & ED Admissions \\
		\hline
		User1 & A1 & 0\\
		\hline
		User1 & A2 & 1\\
		\hline
		User2 & A3 & 0\\
		\hline
	\end{tabular}
	\caption{Number of previous emergency department admissions}
	\label{nopedaTable}
\end{table}

\subsubsection{Admitting Diagnosis}
The admitting diagnosis is obtained from the primary diagnosis code of the claim generated on the admission day. Since this is an ICD code and can have upto ~70000 values, this field is categorized into 18 body system groups namely:
\begin{enumerate}
	\item Infectious and parasitic disease
	\item Neoplasms
	\item Endocrine,nutritional,metabolic,immunity disorders
	\item Blood and blood-forming organs
	\item Mental disorders
	\item Nervous system and sense organs
	\item Circulatory system
	\item Respiratory system
	\item Digestive system
	\item Genitourinary system
	\item Complications of pregnancy, childbirth and the puerperium
	\item Skin and subcutaneous tissue
	\item Musculoskeletal system
	\item Congenital anomalies
	\item Certain conditions originating in the perintal perioud
	\item Symptoms, signs and ill-defined conditions
	\item Injury and poisoning
	\item Factors influencing health status and contact with health services
\end{enumerate}
For each admission in the \tablename{~\ref{admissionTable}}, the admitting diagnosis at CCS level in each admission is shown in \tablename{~\ref{adfeaTable}}.
\begin{table}[!htb]
	\centering
	\begin{tabular}{|c|c|c|}
		\hline
		UserID & ID & Admitting Diagnosis \\
		\hline
		User1 & A1 & Others\\
		\hline
		User1 & A2 & Nervous system and sense organs\\
		\hline
		User2 & A3 & Respiratory system\\
		\hline
	\end{tabular}
	\caption{Admitting Diagnosis for each admission}
	\label{adfeaTable}
\end{table}

\subsubsection{Number of previous hospital visits}
In the medical claims data, hospital visits can be identified by the following CPT Codes 99218-99223 and 99251-99254. The number of previous hospital visits is obtained by taking the count of claims whose service end date is before the service start date of the current admission. For each admission in the \tablename{~\ref{admissionTable}}, the number of previous hospital visits is shown in \tablename{~\ref{nophvTable}}.
\begin{table}[!htb]
	\centering
	\begin{tabular}{|c|c|c|}
		\hline
		UserID & ID & Number of previous\\&& hospital visits \\
		\hline
		User1 & A1 & 0\\
		\hline
		User1 & A2 & 0\\
		\hline
		User2 & A3 & 0\\
		\hline
	\end{tabular}
	\caption{Number of previous hospital visits}
	\label{nophvTable}
\end{table}

\begin{table}[!htb]
	\centering
	\begin{tabular}{|c|c|c|}
		\hline
		UserID & ID & Admission Procedures \\
		\hline
		User1 & A1 & Incision and excision \\&&of CNS\\
		\hline
		User1 & A2 & None\\
		\hline
		User2 & A3 & Gastric bypass and \\ &&volume reduction\\
		\hline
	\end{tabular}
	\caption{Admission Procedures for each admission}
	\label{apfeaTable}
\end{table}

\subsubsection{Admission Procedures}
The procedures taken during an admission are identified through the CPT codes. Using the start and end dates of an admission, the CPT codes from the medical claims generated between these dates are considered. As the number of CPT codes is huge, they are categorized into 242 groups using the Clinical Classification Software (CCS) \cite{ccswebsite}.  The Clinical Classifications Software (CCS) procedure categorization scheme that can be employed in many types of projects analyzing data on procedures. CCS is based on the International Classification of Diseases, a uniform and standardized coding system. The procedure codes are collapsed into a smaller number of clinically meaningful categories that are sometimes more useful for presenting descriptive statistics than the individual codes. For each admission in the \tablename{~\ref{admissionTable}}, the procedures undertaken are shown in \tablename{~\ref{apfeaTable}}.

The extracted predictor variables from the data are combined together and processed for modelling.

%

\section{Data Modelling and Results}

\begin{table*}
	\centering
	\begin{tabular}{|c|c|c|c|c|c|c|c|}
		\hline
		Type & Train AUC & Test AUC & Train & Test & Train  & Test \\
		&&&Specificity&Specificity&Sensitivity&Sensitivity\\
		\hline
		Without &0.716&0.663&0.992&0.992&0.057&0.0591\\
		Feature selection &&&&&&\\
		\hline
		With &0.691&0.659&0.991&0.991&0.0501&0.053\\
		Feature selection &&&&&&\\
		\hline
		PCA Without &0.699&0.655&0.991&0.991&0.0419&0.0419\\
		Feature selection &&&&&&\\
		\hline
		PCA With &0.684&0.660&0.991&0.991&0.0419&0.0419\\
		Feature selection &&&&&&\\
		\hline
		Random Forest&0.85&0.67&0.92&0.90&0.62&0.28\\
		\hline
		SVM&0.66&0.64&0.51&0.50&0.63&0.62\\
		\hline
	\end{tabular}
	\caption{Performance Metrics}
	\label{table:metrics}
\end{table*}

A key objective of this study was to construct predictive models that can predict whether a patient will be readmitted within 30 days, after being discharged from a hospital unit. After extracting the features all the categorical variables are split into multiple binary columns based on the number of levels. Multiple models with response as \textit{Readmission} were built on the dataset using different prediction tools and their performances were compared. The predictive models are based on Logistic Regression, Principal Component Analysis, Random Forest and Support Vector Machines. The dataset for modelling is split into two sets (80\% of the data as training set and 20\% as testing set).

\subsubsection{Logistic Regression}
Two logistic regression models were built one using all the predictor variables (A) and the other using important variables given by log likelihood feature selection (B). Model A has a Train AUC of 0.716 and Test AUC of 0.663 whereas Model B has a Train AUC of 0.691 and Test AUC of 0.659. The metrics of performance are shown in~\tablename{~\ref{table:metrics}}. These results show that the model with all variables has better performance compared to the model with important variables.

\subsubsection{Principal Component Analysis based Regression}
As the modelling ready dataset has very high number of features, Principal Component Analysis(PCA) is performed before and after feature selection on variables. After the PCA, two logistic regression models are built one without feature selection (C) and with feature selection (D).
Model C yielded a Train AUC of 0.699 and a Test AUC of 0.655 whereas Model D yielded a Train AUC of 0.684 and a Test AUC of 0.660. The metrics of performance are shown in~\tablename{~\ref{table:metrics}}.

\subsubsection{Random Forest Classification}
A grid search is performed using Random Forest Machine Learning technique to predict the chance of readmission. For this, the following values of parameters are considered.
\begin{enumerate}
	\item Number of trees (ntree): 500, 1000, 150
	\item Number of variables in Random Sample at each split (mtry): 20, 30, 40, 50
	\item Minimum size of terminal nodes (nodesize): 1, 3, 7, 9
	\item Maximum number of terminal nodes the forest can have (maxnodes): 200, 300
\end{enumerate}
Random Forest models using ten fold cross validation are built using all combinations of the above menitioned parameters. The model with the parameters ntree[500], mtry[50], nodesize[7] and maxnodes[300] gave the best performance. The Train AUC of this model is 0.85 and the Test AUC of this model is 0.67. The ROC plots are shown in the figures~\figurename{\ref{train_roc_rf}} and \figurename{\ref{test_roc_rf}}. The other metrics of performance are shown in~\tablename{~\ref{table:metrics}}. The important features based on gini index from random forest are shown in~\figurename{~\ref{imp_fts_rf}}.

%

\begin{figure}[H]
	\includegraphics[width=0.45\textwidth]{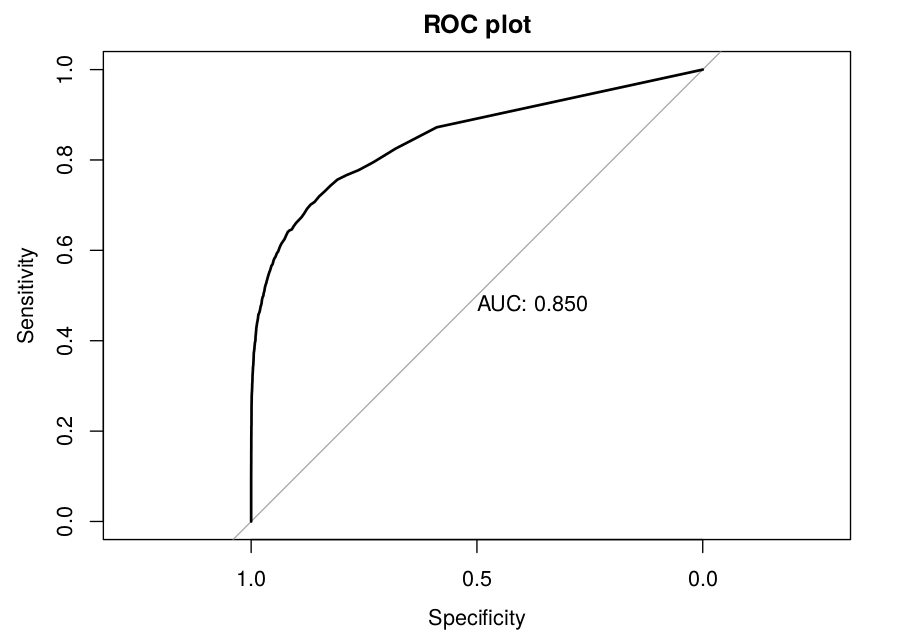}
	\caption{Train ROC Curve for Random Forest}
	\label{train_roc_rf}
\end{figure}

\begin{figure}[H]
	\includegraphics[width=0.45\textwidth]{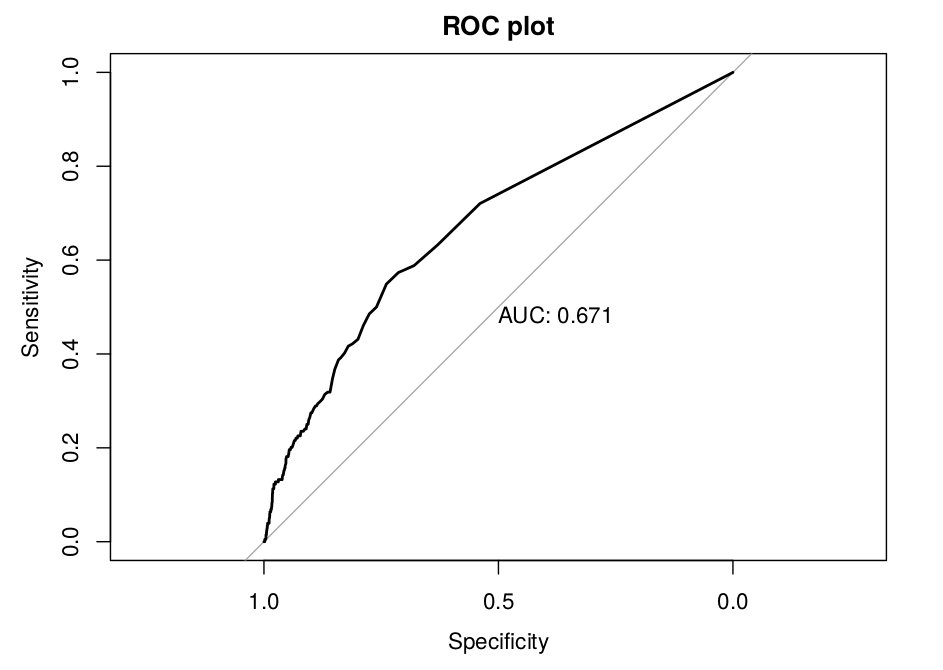}
	\caption{Test ROC Curve for Random Forest}
	\label{test_roc_rf}
\end{figure}

\subsubsection{Support Vector Machine Classification}
Support Vector Machine (SVM) classfication technique is also used to predict the chance of readmission. Several models using ten fold cross validation are built by tuning the cost of constraints violation (C) parameter with Linear kernels. The values of C are mentioned below.
\begin{enumerate}
	\item Kernel: Linear
	\item Cost of constraints violation (C): 0.001, 0.01, 0.05, 0.1, 0.15, 0.2, 0.3, 0.5, 1
\end{enumerate}
The best SVM model had a Train AUC of 0.66 and Test AUC of 0.64. The other metrics of performance are shown in~\tablename{~\ref{table:metrics}}.

\begin{figure}[H]
	\includegraphics[width=0.45\textwidth]{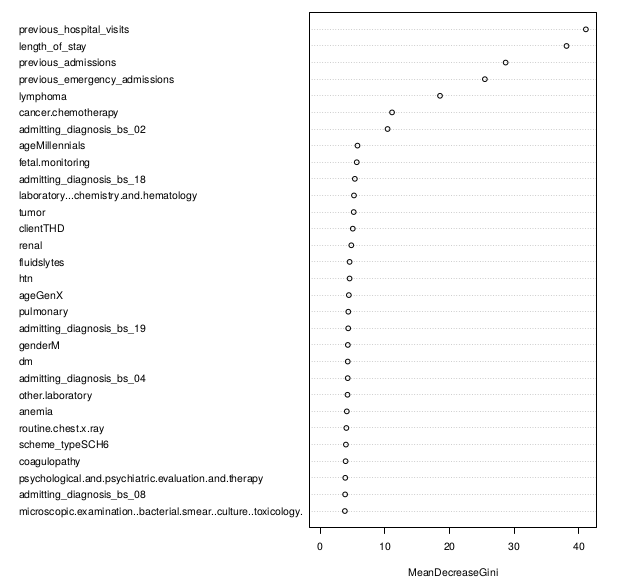}
	\caption{Important Features from Random Forest}
	\label{imp_fts_rf}
\end{figure}



\section{Conclusion}
Reducing the readmission rate is one of the primary actions that can help in achieving a reduction in healthcare expenses. Different strategies can be implemented using the results from predictive modelling. The ability to recognize patients at high risk of readmission is an important step to improve the quality of care. This also helps in targeting interventions to lower the risk of readmission. In this project, we built models to predict all-cause readmissions using medical claims data. Random Forest classification model had the best AUC value. As a part of future work, we aim to build predictive models focussing on specific medical conditions. Pre-index-admission and Post-index-admission data can be used along with the admission data to understand the crucial causes behind a readmission. 

\bibliographystyle{plain}
\bibliography{HospitalReadmissionReferences}

\end{document}